\documentclass{article} %
\usepackage[preprint]{colm2026_conference}

\usepackage{microtype}
\usepackage{hyperref}
\usepackage{url}
\usepackage{booktabs}

\usepackage{graphicx}
\usepackage{multirow}
\usepackage[dvipsnames]{xcolor}
\usepackage[most]{tcolorbox}
\usepackage[T1]{fontenc}
\usepackage[utf8]{inputenc}
\usepackage{subcaption}
\newtcolorbox{block}[1][]{
    colback=gray!10!white, colframe=gray!80!black,        
    fonttitle=\bfseries,        %
    title=#1,                   
    boxrule=0.5mm,              %
    rounded corners,            %
    before skip=10pt,           %
    after skip=10pt,            %
}

\usepackage{lineno}

\definecolor{darkblue}{rgb}{0, 0, 0.5}
\hypersetup{colorlinks=true, citecolor=darkblue, linkcolor=darkblue, urlcolor=darkblue}

\title{Adapting English Quality Classifiers for Multilingual LLM Pretraining Data Selection}

\author{Vinko Sabolčec, Bettina Messmer, Yassine Turki \& Martin Jaggi\\
EPFL\\
\texttt{firstname.lastname@epfl.ch} \\
}

\begin{document}

\ifcolmsubmission
\linenumbers
\fi

\maketitle

\begin{abstract}

Recent advances in large language model (LLM) pretraining highlight the role of high-quality training data in improving performance.
While model-based filtering has proven effective in selecting high-quality subsets from web-scale corpora, especially for high-resource languages, low-resource languages face challenges due to limited availability of annotated data.
This work explores extending quality filtering to over 100 languages by proposing a multilingual adaptation approach that converts an existing English quality classifier into a multilingual variant.
Our approach proposes training a small multi-layer perceptron on top of Transformer encoder-only model embeddings, using multilingual text as input and scores obtained from English classifiers applied to machine-translated text as labels.
Our 1B, 3B and 8B scale experiments show that our approach maintains the downstream LLM benchmark performance of existing multilingual model-based filtering baselines, without harming regional and cultural knowledge benchmarks.
To further evaluate cross-lingual generalization, we compare classifier scores of high-quality synthetic data and web samples, and the correlation of classifier scores with LLM-based ones, revealing that the classifier can learn the scoring criteria of its original English variant, even for languages not included in its training data.

\end{abstract}

\section{Introduction}

Training of large language models (LLMs) is increasingly compute-intensive, with a significant portion of resources allocated to pretraining on web-scale corpora. Consequently, data selection presents an important axis for improving downstream performance while minimizing training costs. Recent research has demonstrated that curating high-quality subsets from web corpora can match or even surpass the performance of models trained on larger, less refined datasets~\citep{penedo2023refinedwebdatasetfalconllm,penedo2024finewebdatasetsdecantingweb,li2025datacomplmsearchgenerationtraining,messmer2026enhancingmultilingualllmpretraining,chen2026muratinghighqualitydata}.

Data curation for pretraining has evolved from heuristic filtering and deduplication~\citep{raffel2023exploringlimitstransferlearning,rae2022scalinglanguagemodelsmethods,penedo2023refinedwebdatasetfalconllm,penedo2024finewebdatasetsdecantingweb} toward model-based quality selection. FineWeb-edu~\citep{penedo2024finewebdatasetsdecantingweb} and DCLM~\citep{li2025datacomplmsearchgenerationtraining} demonstrated that selecting high-quality subsets from web corpora can reduce the token budget by up to 10 times to reach baseline performance, while exceeding it on same token budgets. Subsequent work extended this paradigm to multilingual settings. Namely, FineWeb2-HQ~\citep{messmer2026enhancingmultilingualllmpretraining} trained per-language quality classifiers across 20 languages and MuRating~\citep{chen2026muratinghighqualitydata} leveraged multiple English quality classifiers and machine translations to train a quality classifier across 17 languages. However, these methods either depend on language-specific training data or have not been applied beyond high-resource languages. Recent work by~\citet{turki2026crosslingual} expanded the FineWeb2-HQ approach to over 100 languages and analyzed cross-lingual generalization, however, still relying on language-specific data.

This raises a fundamental challenge: how can we scale high-quality data curation to long-tail languages without relying on language-specific annotated data? We propose decoupling quality signals from the underlying language. By projecting these signals across languages rather than relearning them for each one, we provide a scalable, language-agnostic path for data curation.

Our work proposes a simple pipeline that leverages existing English quality classifiers and uses them to train multilingual variants. Specifically, inspired by the FineWeb2-HQ-like approaches~\citep{messmer2026enhancingmultilingualllmpretraining,turki2026crosslingual} and MuRating~\citep{chen2026muratinghighqualitydata}, we propose a multilingual model-based filtering approach that trains a small multi-layer perceptron (MLP) on top of Transformer~\citep{vaswani2023attentionneed} encoder-only model embeddings, using multilingual text embeddings as inputs and scores from English classifiers applied to machine-translated text as labels. This design enables the model to inherit quality criteria from English classifiers while adapting to other languages, reducing reliance on language-specific training data and expanding the language coverage using the embedding model.

To evaluate our approach, we conduct experiments across 1B, 3B, and 8B parameter models. Our results show that our approach maintains downstream LLM benchmark performance comparable to existing multilingual filtering baselines, while preserving regional and cultural knowledge. Further analysis of cross-lingual generalization reveals that the classifier learns the scoring criteria of its original English variant, even for languages not included in the training set. This suggests a promising path toward scalable, language-agnostic data curation that leverages existing work in the high-resource space.

In summary, our contributions are:
\begin{itemize}
    \item We propose an approach that trains multilingual variants of existing English quality classifiers without per-language annotated data. Furthermore, we release the weights of our multilingually adapted FineWeb-edu and DCLM quality classifiers\footnote{\href{https://huggingface.co/epfml/FineWeb2-HQ-PlusPlus-Classifier}{huggingface.co/epfml/FineWeb2-HQ-PlusPlus-Classifier}}.
    \item We empirically evaluate our quality classifiers by pretraining 1B, 3B, and 8B parameter LLMs and compare them to existing baselines, demonstrating comparable downstream benchmark performance. Additionally, we show that the multilingual adaptation process does not inherently introduce regional or cultural biases on downstream benchmarks.
    \item We analyze cross-lingual generalization properties of our quality classifiers, revealing that the classifiers learn the scoring criteria of the original English variant even for languages not seen during training.
\end{itemize}

\section{Related Work}

\paragraph{Pretraining Data Curation.} Early efforts such as CCNet~\citep{wenzek2019ccnetextractinghighquality}, C4~\citep{raffel2023exploringlimitstransferlearning}, and Gopher~\citep{rae2022scalinglanguagemodelsmethods} established pipelines including language identification, heuristic-based filtering, deduplication, and language model perplexity-based filtering to remove low-quality web content. Building on these, RefinedWeb~\citep{penedo2023refinedwebdatasetfalconllm} and FineWeb~\citep{penedo2024finewebdatasetsdecantingweb} introduced more systematic pipelines based on language identification, heuristic-based filtering, and deduplication, resulting in datasets that outperform larger, less curated corpora on downstream benchmarks. Multilingual curation has followed a similar trajectory~\citep{AbadjiOrtizSuarezRomaryetal.2021,abadji-etal-2022-towards,laurencon2023bigsciencerootscorpus16tb,nguyen2023culturaxcleanedenormousmultilingual,degibert2024newmassivemultilingualdataset,burchell2025expandedmassivemultilingualdataset,oepen2025hplt30largescalemultilingual}, with a notable example being FineWeb~2~\citep{penedo2025fineweb2pipelinescale}, which extended such pipelines to over 1800 language-script pairs, demonstrating that quality-oriented heuristics can be adapted beyond English at a significant scale.

\paragraph{Model-based Quality Filtering.} A subsequent line of work moved beyond heuristics by training models to assess text quality. FineWeb-edu~\citep{penedo2024finewebdatasetsdecantingweb} used LLM-based annotations to score web content according to its educational value, while DCLM~\citep{li2025datacomplmsearchgenerationtraining} trained a lightweight FastText~\citep{joulin2016bagtricksefficienttext} binary classifier to distinguish instruction-like text from web data. Both approaches demonstrated that selecting a small, high-quality subset of training data can match or exceed the downstream performance of models trained on an order of magnitude larger unfiltered corpora. Despite their effectiveness, these methods were developed and evaluated exclusively for English, leaving the model-based filtering paradigm unexplored in multilingual settings.

\paragraph{Multilingual Model-based Quality Filtering.} Similar to the DCLM approach, FineWeb2-HQ~\citep{messmer2026enhancingmultilingualllmpretraining} trained per-language quality classifiers for 20 languages, achieving filtering quality comparable to English-only methods and showing improvement in multilingual LLM pretraining performance. Concurrently, MuRating~\citep{chen2026muratinghighqualitydata} further improved performance, leveraging machine translation and aggregated English classifier scores to train a multilingual quality classifier covering 17 languages. Recently, \citet{turki2026crosslingual} expanded FineWeb2-HQ to a single multilingual classifier supporting over 100 languages, with an exploration of sampling strategies and cross-lingual generalization. Despite this, current methodologies are either reliant on the availability of language-specific high-quality training data or have not been evaluated beyond high-resource languages. Similar to MuRating, our work reuses English classifiers and extends them to arbitrary languages through multilingual embeddings and machine-translated text. Importantly, we analyze cross-lingual generalization of this approach on high-resource and low-resource languages both in the classifier training distribution and out the training distribution. While we do not directly compare to MuRating since their code, classifier weights, and data are not publicly available, we note that our findings on cross-lingual generalization offer a promising path to extending its high performance beyond the original 17 languages.

\paragraph{Multilingual Embedding Models.} In data curation, embedding models can effectively capture quality signals necessary for multilingual quality filtering~\citep{messmer2026enhancingmultilingualllmpretraining,turki2026crosslingual}. While early multilingual Transformer~\citep{vaswani2023attentionneed} encoder-only models like mBERT~\citep{devlin2018bert} and XLM-RoBERTa~\citep{conneau2020unsupervisedcrosslingualrepresentationlearning} supported roughly 100 languages, the recent mmBERT~\citep{marone2025mmbertmodernmultilingualencoder} has expanded this to over 1800 language-script pairs. Our approach leverages these embeddings by training a lightweight MLP on top of them.

\section{Method}

We present our approach for extending English quality classifiers to multilingual settings. Our pipeline consists of three stages: translating multilingual documents into English, applying existing English classifiers to generate quality scores and obtaining the training data by generating text embeddings, and training classifiers using the obtained data.

We base our method on the English FineWeb and the multilingual FineWeb~2 datasets due to their strong baseline performance and their support for over 1800 language-script pairs. Throughout the paper, \textit{top N languages} refers to the language-script pairs present in the FineWeb~2 dataset ranked by byte count in the training split.

\paragraph{Multilingual Document Translation.} To apply English quality classifiers to the multilingual FineWeb 2 dataset, we machine-translate a random 0.1\% sample of the top 100 language data into English. The reason for our choice of translating multilingual text into English, as opposed to translating quality-scored English samples into multilingual text, is twofold. Firstly, we aim to preserve the distribution and the nuances of the multilingual web to reduce the English homogeinity of the training data, and, secondly, we observed that the translation models we used during our initial trials were better at this translation direction. For the translation process, we use the \texttt{Qwen3-32B}\footnote{\url{https://huggingface.co/Qwen/Qwen3-32B}}~\citep{yang2025qwen3technicalreport} model, which supports 119 languages and dialects, with the datatrove~\citep{penedo2024datatrove} and vLLM~\citep{kwon2023efficient} libraries, and we disable the model's reasoning outputs. We design our translation prompt to preserve the original text's characteristics, including its formatting, style, and even organic errors. The full prompt and a translated example are available in Appendix~\ref{appendix:translation}. This results in $\sim$5M translated documents, for which we use approximately 750 H100 GPU hours.

\paragraph{Classifier Training Data.} We score the English translations of multilingual data using two widely used English quality classifiers: DCLM\footnote{\url{https://huggingface.co/mlfoundations/fasttext-oh-eli5}} and FineWeb-edu\footnote{\url{https://huggingface.co/HuggingFaceFW/fineweb-edu-classifier}}. Additionally, we score a subset of English FineWeb data for use during classifier training. For training the multilingual classifiers, we follow an approach inspired by FineWeb2-HQ~\citep{messmer2026enhancingmultilingualllmpretraining} and the multilingual extension of \citet{turki2026crosslingual}, training a lightweight MLP on top of frozen Transformer encoder-only embeddings. During training, the MLP takes embeddings of the original multilingual text as inputs and predicts the quality scores obtained from the English classifiers. This frozen-embedding approach minimizes computational overhead and facilitates embedding reuse across tasks, though the framework remains compatible with full fine-tuning. We experiment with two multilingual embedding models: XLM-RoBERTa~\citep{conneau2020unsupervisedcrosslingualrepresentationlearning}, which supports 100 languages, and mmBERT~\citep{marone2025mmbertmodernmultilingualencoder}, which supports over 1800 language-script pairs. The final training set consists of $\sim$6M samples, of which $\sim$1M are English and $\sim$5M are multilingual.

\paragraph{Classifier Training.} 
The MLP consists of two hidden layers of dimension 3072 with ReLU activations and a single scalar output. We train the classifier using the AdamW~\citep{loshchilov2019decoupledweightdecayregularization} optimizer with a cosine learning rate schedule with 10\% linear warmup, peaking at 3e-4, for 6 epochs with a batch size of 1024.
The loss function is adapted to the output format of each classifier. For FineWeb-edu, which produces a scalar score in [0, 5], we apply L1 loss directly on the MLP output. For DCLM, which outputs a probability in [0, 1], we compute the L1 loss on the logit transformation, $\ln \left(\frac{p}{1-p}\right)$, of the probability scores\footnote{We clip the probabilities to $[\epsilon, 1-\epsilon]$ with $\epsilon = \texttt{1e-6}$ to avoid numerical instabilities, implemented in PyTorch as \texttt{torch.special.logit(probs, eps=1e-6)}.}. We show the score distribution of the original classifiers and the multilingual adaptations, score error distribution, and the correlation between between their scores on English FineWeb data in Appendix~\ref{appendix:eng_multiling_classifier_comparison}. Unless stated otherwise, in our experiments, we train the classifier models on data of all languages we processed (English and top 100 languages).

\section{Results and Discussion}

In this section, we detail our experimental setup for LLM pretraining and cross-lingual generalization analysis, which we use to validate our approach. We then present our results on identifying the differences between combinations of different English base classifiers and embedding models, the impact of language quantity during classifier training, performance against other baselines, including a regional and cultural bias analysis, and analyze cross-lingual generalization of our classifiers.

\subsection{Experimental Setup}

\paragraph{LLM Pretraining.}  To evaluate our classifiers, we pretrain 1B, 3B, and 8B parameter LLMs on data filtered using quality classifier models. Following prior work \citep{li2025datacomplmsearchgenerationtraining, messmer2026enhancingmultilingualllmpretraining}, we select the top 10\% highest-quality data per language from FineWeb and FineWeb 2 using the datatrove~\citep{penedo2024datatrove} library. For multilingual model training, we preserve the original language distribution and rehydrate FineWeb 2 data according to hyperparameters by \citet{penedo2025fineweb2pipelinescale}. We use the Megatron-LM~\citep{megatron-lm} framework and the Apertus~\citep{apertus2025apertusdemocratizingopencompliant} architecture. All runs use a 4096 sequence length, the AdEMAMix optimizer \citep{pagliardini2024ademamixoptimizerbetterfaster}, and a Warmup-Stable-Decay scheduler \citep{hu2024minicpmunveilingpotentialsmall, hagele2024scalinglawscomputeoptimaltraining}. Specific training configurations vary across model sizes to balance our computational constraints and signal obtained from training the models, and are available in Appendix~\ref{appendix:pretrain_configs}. For our experiments, we use a compute cluster with each node containing 4 H100 GPUs, and use 21 nodes for 1B parameter models, and 64 nodes for 3B and 8B models. In total, we use approximately 33k H100 GPU hours for training our models.

\paragraph{LLM Evaluation.} We evaluate the models using the lm-eval-harness~\citep{eval-harness} library, using a benchmark suite targeting language understanding, reasoning, and global and local knowledge acquisition. Our benchmark suite consists of the lite subset of Global-MMLU (GMMLU;~\citealt{singh2025globalmmluunderstandingaddressing}), INCLUDE~\citep{romanou2024includeevaluatingmultilinguallanguage}, ARC Easy and Challenge~\citep{clark2018thinksolvedquestionanswering}, multilingual translation of ARC (mARC;~\citealt{lai2023okapiinstructiontunedlargelanguage}), HellaSwag (HS;~\citealt{zellers2019hellaswagmachinereallyfinish}), multilingual translation of HellaSwag (mHS;~\citealt{lai2023okapiinstructiontunedlargelanguage}), XNLI~\citep{conneau2018xnlievaluatingcrosslingualsentence}, OpenBookQA (OBQA;~\citealt{mihaylov2018suitarmorconductelectricity}), XWinoGrad (XWG;~\citealt{muennighoff2023crosslingualgeneralizationmultitaskfinetuning,tikhonov2021itsheadsusingattention}), and the easy subset of CulturalBench~\citep{chiu2025culturalbenchrobustdiversechallenging}. Inspired by FineTasks~\citep{kydlicek2024finetasksmultilingualtasks}, we select our benchmarks based on our initial 1B model pretraining runs to prioritize consistent ordering, non-random signal and monotonic increase during training to maximize the obtained signal in our experiments. Furthermore, for Global-MMLU and INCLUDE, we use the cloze formulation since the multiple-choice formulation results in random-guess accuracy for smaller models~\citep{kydlicek2024finetasksmultilingualtasks}. Similarly, we use CulturalBench only for 8B parameter models, since we observe volatile performance at smaller scales. When reporting results, we use the normalized accuracy metric, and in case of multilingual benchmarks, we average the per-language results of the benchmark.

\paragraph{Dataset Baselines.} We evaluate our multilingually adapted classifiers against several baselines: FineWeb and FineWeb~2 (FW+FW2;~\citealt{penedo2024finewebdatasetsdecantingweb,penedo2025fineweb2pipelinescale}), FineWeb-HQ and FineWeb2-HQ (FWHQ+FW2HQ;~\citealt{messmer2026enhancingmultilingualllmpretraining}), and the expanded multilingual FineWeb2-HQ classifier (FWHQ+FW2HQ$^{+}$;~\citealt{turki2026crosslingual}). Since FineWeb2-HQ only supports 20 languages, when training on all languages, we sample from FineWeb~2 for the remaining languages, maintaining the same per-language distribution found in the full FineWeb~2 corpus. For all other model-based filtering approaches, we apply the quality classifiers to all training and evaluation languages.

\paragraph{Cross-lingual Generalization Analysis.} To assess cross-lingual generalization, we categorize languages into three tiers based on data availability and training feasibility:
\begin{enumerate}
    \item \textbf{English + Top 20 Languages.} We follow standard pretraining and evaluation procedure consistent with prior work~\citep{messmer2026enhancingmultilingualllmpretraining,turki2026crosslingual}, as described above.
    \item \textbf{Top 21 -- 100 Languages.} Due to limited available benchmarks and low data distribution coverage in small-scale training runs, we evaluate generalization by comparing classifier scores on higher-quality synthetic encyclopedia-like data and lower-quality web samples from FineWeb and FineWeb 2. We generate the synthetic encyclopedia-like data using the \texttt{Qwen3.5-35B-A3B}\footnote{\url{https://huggingface.co/Qwen/Qwen3.5-35B-A3B}} model for 8 languages and detail the process in Appendix~\ref{appendix:synth_encyclopedia}.
    \item \textbf{Beyond Top 100.} Due to limited available models with generation capabilities for these languages, we use the LLM-as-a-judge approach. We use the \texttt{Qwen3.5-35B-A3B}, \texttt{gemma-3-27b-it}\footnote{\url{https://huggingface.co/google/gemma-3-27b-it}}, and \texttt{gpt-oss-120b}\footnote{\url{https://huggingface.co/openai/gpt-oss-120b}} models and correlate their assigned scores with quality classifier scores on web samples. For the LLM judges, we use a FineWeb-edu-like prompt, which we show in Appendix~\ref{appendix:llm_judge} with a judged example.
\end{enumerate}

\subsection{Experimental Results}

\paragraph{Which Classifier and Embedding Model Perform the Best?} To identify the most effective combination of base classifiers paired with different embedding models, we evaluate the FineWeb-edu (mFW-edu) and DCLM-based (mDCLM) classifiers using XLM-RoBERTa (XLM-R) and mmBERT embeddings. We present the downstream benchmark results of pretrained LLMs in Table~\ref{tab:classifier_vs_embeding}. The results indicate that FineWeb-edu and DCLM yield comparable performance across both embedding models, highlighting that both embedding models are able to capture relevant quality signals at this scale. Furthermore, similar performance between the mFW-edu and mDCLM classifiers is consistent with prior work~\citep{li2025datacomplmsearchgenerationtraining}. Given the similar downstream benchmark performance, in the rest of our experiments, we will focus only on the models based on the mmBERT model embeddings.

\begin{table}[!htb]
    \centering
\resizebox{\textwidth}{!}{
\begin{tabular}{llcccccccccc}
\toprule[2pt]
 & & \textbf{GMMLU$_c$} & \textbf{INCLUDE$_c$} & \textbf{ARC} & \textbf{mARC} & \textbf{HS} & \textbf{mHS} & \textbf{XNLI} & \textbf{OBQA} & \textbf{XWG} & \textbf{Avg.} \\
\textbf{Classifier}  & \textbf{Emb. Model} &  &  &  &  &  &  &  &  &  &  \\

\midrule[2pt]
\multicolumn{10}{l}{\textit{1B Model, 100B Tokens from Scratch, Training and Evaluation on English + Top 20 Languages}} \\
\midrule

\multirow{2}{*}{mFW-edu} & XLM-R & 0.2814 & 0.2944 & 0.4812 & 0.2936 & 0.5412 & 0.4034 & 0.4161 & \textbf{0.3800} & 0.6046 & 0.4106 \\
 & mmBERT & \textbf{0.2989} & 0.2913 & \textbf{0.4897} & \textbf{0.2937} & 0.5469 & 0.4026 & 0.4121 & 0.3740 & 0.6180 & 0.4141 \\
\multirow{2}{*}{mDCLM} & XLM-R & 0.2845 & \textbf{0.2965} & 0.4802 & 0.2815 & \textbf{0.5651} & \textbf{0.4080} & \textbf{0.4227} & 0.3640 & 0.6277 & \textbf{0.4145} \\
 & mmBERT & 0.2848 & 0.2938 & 0.4554 & 0.2744 & 0.5454 & 0.4012 & 0.4142 & 0.3780 & \textbf{0.6314} & 0.4087 \\

\bottomrule
\end{tabular}
}
    \caption{Benchmark performance of 1B parameter multilingual LLMs trained on 100B tokens. We compare models trained on data selected using our multilingually adapted mFW-edu and mDCLM quality classifiers, paired with XLM-RoBERTa and mmBERT embeddings. We retain the top 10\% of documents based on the classifier scores.}
    \label{tab:classifier_vs_embeding}
\end{table}

\begin{block}
\textbf{Takeaway.} The choice between mDCLM or mFW-edu classifiers and XLM-RoBERTa or mmBERT embedding models results in only modest variations in downstream performance.
\end{block}

\begin{figure}[!htb]
    \centering
    \includegraphics[width=0.65\linewidth]{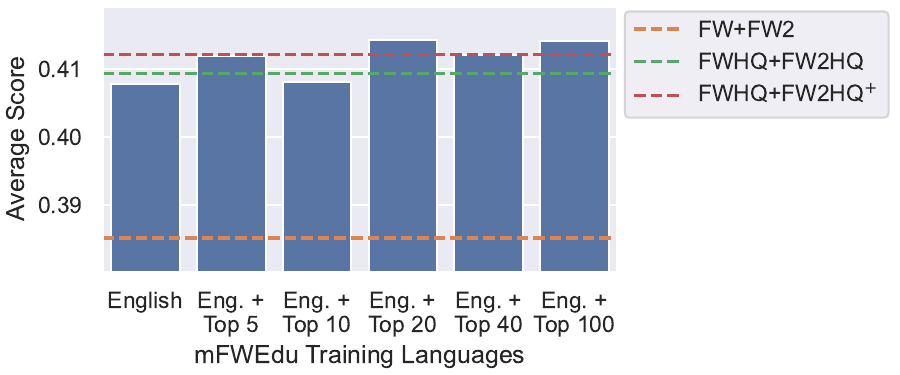}
    \caption{Benchmark performance of 1B parameter multilingual LLMs trained on 100B tokens. We compare models trained on data selected using our multilingually adapted mFW-edu classifier with mmBERT embeddings trained on a varying amount of languages, and the FW+FW2, FWHQ+FW2HQ, and FWHQ+FW2HQ$^{+}$ baselines. We retain the top 10\% of documents based on the classifier scores, except for FW+FW2 which is not model-filtered.}
    \label{fig:ablation_classifier_languages}
\end{figure}

\paragraph{How Does the Number of Classifier Training Languages Affect Performance?} To determine how classifier language quantity influences downstream LLM performance, we evaluate 1B parameter models trained on 100B tokens from English and the top 20 languages in FineWeb~2. Figure \ref{fig:ablation_classifier_languages} compares the mFW-edu classifier with mmBERT embeddings, trained on varying number of languages, against our baselines. Additionally, we include the table with all evaluation benchmark results in Appendix~\ref{appendix:varying_languages_classifier}. We find that, even when trained exclusively on English, the mFW-edu classifier significantly outperforms the FW+FW2 baseline and remains competitive with other model-based approaches. While performance remains stable when training on the top 5 or 10 languages, we observe a slight gain when the classifier is trained on the full set of languages used during LLM training and evaluation.

\begin{block}
\textbf{Takeaway.} Classifiers can achieve near-peak performance even when trained on a subset of target languages.
\end{block}

\paragraph{How Does the Approach Compare to Existing Baselines?} We compare our multilingually adapted classifiers, mFW-edu and mDCLM with mmBERT embeddings, against our baselines across 1B, 3B, and 8B scales. We present the results in Table~\ref{tab:baseline_comparison}. Our results confirm a consistent performance hierarchy: FWHQ+FW2HQ model-based filtered dataset outperforms the heuristic-based filtered FW+FW2, while the FWHQ+FW2HQ$^{+}$ achieves the highest baseline performance. Notably, our adapted classifiers, mFW-edu and mDCLM, perform comparably to this strongest baseline. Specifically, mFW-edu is the best performing at the 1B scale, while mDCLM performs the best at 3B and 8B scales on average. These findings align with prior work in the English~\citep{li2025datacomplmsearchgenerationtraining} and the multilingual space~\citep{messmer2026enhancingmultilingualllmpretraining,turki2026crosslingual}.

\begin{table}[!htb]
    \centering
\resizebox{\textwidth}{!}{

\begin{tabular}{lcccccccccc}
\toprule[2pt]
& \textbf{GMMLU$_c$} & \textbf{INCLUDE$_c$} & \textbf{ARC} & \textbf{mARC} & \textbf{HS} & \textbf{mHS} & \textbf{XNLI} & \textbf{OBQA} & \textbf{XWG} & \textbf{Avg.} \\
\textbf{Model} &  &  &  &  &  &  &  &  &  & \\

\midrule[2pt]
\multicolumn{11}{l}{\textit{1B Model, 100B Tokens from Scratch, Training and Evaluation on English + Top 20 Languages}} \\
\midrule

FW+FW2 & 0.2743 & 0.2817 & 0.3880 & 0.2586 & 0.5434 & 0.3887 & 0.4086 & 0.3180 & 0.6051 & 0.3852 \\
FWHQ+FW2HQ & 0.2893 & 0.2870 & 0.4461 & 0.2747 & 0.5641 & 0.4035 & \textbf{0.4214} & 0.3680 & 0.6308 & 0.4094 \\
FWHQ+FW2HQ$^{+}$ & 0.2952 & \textbf{0.2959} & 0.4505 & 0.2816 & \textbf{0.5762} & \textbf{0.4120} & 0.4116 & 0.3640 & 0.6221 & 0.4121 \\
mFW-edu (Ours) & \textbf{0.2989} & 0.2913 & \textbf{0.4897} & \textbf{0.2937} & 0.5469 & 0.4026 & 0.4121 & 0.3740 & 0.6180 & \textbf{0.4141} \\
mDCLM (Ours) & 0.2848 & 0.2938 & 0.4554 & 0.2744 & 0.5454 & 0.4012 & 0.4142 & \textbf{0.3780} & \textbf{0.6314} & 0.4087 \\

\midrule[2pt]
\multicolumn{11}{l}{\textit{3B Model, 50B Token Cooldown from 200B Token Checkpoint, Training and Evaluation on All Available Languages}} \\
\midrule

Pre-cooldown & 0.2885 & 0.2816 & 0.4997 & 0.2668 & 0.6967 & 0.3753 & 0.4029 & 0.3800 & 0.6410 & 0.4258 \\
FW+FW2 & 0.2927 & 0.3012 & 0.4953 & 0.2792 & 0.6922 & 0.4149 & 0.4192 & 0.4000 & 0.6656 & 0.4400 \\
FWHQ+FW2HQ+FW2 & 0.2995 & 0.3026 & 0.5300 & 0.2893 & \textbf{0.6982} & \textbf{0.4223} & 0.4165 & 0.4060 & 0.6816 & 0.4496 \\
FWHQ+FW2HQ$^{+}$ & 0.3012 & \textbf{0.3088} & \textbf{0.5522} & \textbf{0.2979} & 0.6850 & 0.4178 & \textbf{0.4215} & 0.4080 & 0.6628 & 0.4506 \\
mFW-edu (Ours) & 0.2987 & 0.3083 & 0.5520 & 0.2979 & 0.6857 & 0.4180 & 0.4205 & 0.4100 & 0.6594 & 0.4501 \\
mDCLM (Ours) & \textbf{0.3032} & 0.3036 & 0.5345 & 0.2952 & 0.6903 & 0.4205 & 0.4206 & \textbf{0.4180} & \textbf{0.6891} & \textbf{0.4528} \\

\midrule[2pt]
\multicolumn{11}{l}{\textit{8B Model, 63B Token Cooldown from 9.5T Token Checkpoint, Training and Evaluation on All Available Languages}} \\
\midrule

Pre-cooldown & 0.3657 & 0.3774 & 0.6457 & 0.3800 & \textbf{0.7898} & 0.5347 & 0.4420 & 0.4360 & 0.7436 & 0.5239 \\
FW+FW2 & 0.3708 & 0.3804 & 0.6544 & 0.3850 & 0.7883 & 0.5374 & 0.4431 & 0.4420 & 0.7550 & 0.5285 \\
FWHQ+FW2HQ+FW2 & 0.3823 & 0.3781 & 0.6466 & 0.3869 & 0.7872 & 0.5404 & 0.4474 & 0.4460 & 0.7565 & 0.5302 \\
FWHQ+FW2HQ$^{+}$ & 0.3790 & 0.3860 & 0.6597 & 0.3932 & 0.7889 & \textbf{0.5423} & \textbf{0.4496} & \textbf{0.4520} & 0.7518 & 0.5336 \\
mFW-edu (Ours) & \textbf{0.3843} & 0.3820 & 0.6585 & 0.3910 & 0.7868 & 0.5404 & 0.4431 & 0.4400 & \textbf{0.7613} & 0.5319 \\
mDCLM (Ours) & 0.3815 & \textbf{0.3863} & \textbf{0.6646} & \textbf{0.3956} & 0.7861 & 0.5391 & 0.4443 & \textbf{0.4520} & 0.7602 & \textbf{0.5344} \\

\bottomrule
\end{tabular}

}
    \caption{Benchmark performance of 1B, 3B, and 8B parameter multilingual LLMs. We compare models trained on data selected using our multilingually adapted mFW-edu and mDCLM quality classifiers paired with mmBERT embeddings, and the FW+FW2, FWHQ+FW2HQ, and FWHQ+FW2HQ$^{+}$ baselines. We retain the top 10\% of documents based on the classifier scores, except for FW+FW2 which is not model-filtered.}
    \label{tab:baseline_comparison}
\end{table}

\begin{block}
\textbf{Takeaway.} Adapting English quality classifiers for multilingual LLM pretraining data selection results in downstream performance comparable to existing baselines, with performance being consistent with prior English and multilingual work on model-based quality filtering.
\end{block}

\paragraph{Does the Multilingual Adaptation Process Introduce Regional or Cultural Biases?} To ensure our multilingual adaptation doesn't introduce English-centric biases, we evaluate downstream LLM performance on regional (INCLUDE;~\citealt{romanou2024includeevaluatingmultilinguallanguage}) and cultural (CulturalBench;~\citealt{chiu2025culturalbenchrobustdiversechallenging}) knowledge benchmarks. To isolate adaptation-related biases from those inherent to the classifier, we analyze the FineWeb-edu-based mFW-edu classifier, which is based on a culture-neutral prompt targeting educational content, and compare it against the strongest native multilingual baseline, FWHQ+FW2HQ$^{+}$. To test the limits of cross-lingual transfer, we also include the mFW-edu$_{\text{Eng}}$ classifier, a version of the classifier trained using only English data. Tables \ref{tab:include_1B_30BT_monoling} and \ref{tab:culturalbench_8B_multiling} summarize the performance of monolingual 1B models across four languages on INCLUDE and multilingual 8B models on CulturalBench, respectively. The results indicate that the adaptation process does not introduce a bias. mFW-edu performs comparably with the native multilingual baseline across both benchmarks. While the English-only adapted classifier (mFW-edu$_{\text{Eng}}$) maintains parity with other approaches at the 1B scale, its performance drops at the 8B scale on the CulturalBench benchmark.

\begin{table}[!htb]
    \small
    \centering
\resizebox{0.8\textwidth}{!}{
\begin{tabular}{lccccc}
\toprule[2pt]
 & \textbf{Chinese} & \textbf{Japanese} & \textbf{French} & \textbf{Portuguese} & \textbf{Avg.} \\
\textbf{Model} &  &  &  &  &  \\

\midrule[2pt]
\multicolumn{6}{l}{\textit{1B Model, 30B Tokens From Scratch, Monolingual Training and Evaluation}} \\
\midrule

FWHQ+FW2HQ$^{+}$ & \textbf{0.3578} & 0.3453 & 0.3938 & 0.3140 & 0.3527 \\
mFW-edu (Ours) & 0.3358 & \textbf{0.3653} & \textbf{0.4153} & 0.3230 & \textbf{0.3598} \\
mFW-edu$_\text{Eng}$ (Ours) & 0.3541 & 0.3453 & 0.3771 & \textbf{0.3339} & 0.3526 \\

\bottomrule
\end{tabular}
}
    \caption{Benchmark performance of 1B monolingual Chinese, Japanese, French and Portuguese LLMs on the INCLUDE benchmark. We compare models trained on data selected using our multilingually adapted mFW-edu quality classifier paired with mmBERT embeddings trained on all data (mFW-edu), trained on English (mFW-edu$_{\text{Eng}}$), and the FWHQ+FW2HQ$^{+}$ baseline. We retain the top 10\% of documents based on the classifier scores.}
    \label{tab:include_1B_30BT_monoling}
\end{table}

\begin{table}[!htb]
    \small
    \centering
\resizebox{0.8\textwidth}{!}{
\begin{tabular}{lccccc}
\toprule[2pt]
 & \textbf{China} & \textbf{Japan} & \textbf{France} & \textbf{Brazil} & \textbf{Avg. (All Subsets)} \\
\textbf{Model} &  &  &  &  & \\

\midrule[2pt]
\multicolumn{6}{l}{\textit{8B Model, 63B Token Cooldown from 9.5T Token Checkpoint, Training and}} \\
\multicolumn{6}{l}{\textit{Evaluation on All Available Languages}} \\
\midrule

Pre-cooldown & 0.4915 & 0.6792 & \textbf{0.6429} & \textbf{0.4000} & 0.5764 \\
FWHQ+FW2HQ$^{+}$ & 0.5424 & \textbf{0.7925} & \textbf{0.6429} & 0.3600 & 0.6143 \\
mFW-edu (Ours) & \textbf{0.5763} & 0.7170 & 0.5714 & \textbf{0.4000} & \textbf{0.6164} \\
mFW-edu$_{\text{Eng}}$ (Ours) & 0.5254 & 0.7358 & 0.5714 & 0.3600 & 0.5931 \\

\bottomrule
\end{tabular}
}
    \caption{Benchmark performance of 8B multilingual LLMs on the easy subset of the CulturalBench benchmark. We compare models trained on data selected using our multilingually adapted mFW-edu quality classifier paired with mmBERT embeddings trained on all data (mFW-edu), trained on English (mFW-edu$_{\text{Eng}}$), and the FWHQ+FW2HQ$^{+}$ baseline. We retain the top 10\% of documents based on the classifier scores.}
    \label{tab:culturalbench_8B_multiling}
\end{table}

\begin{block}
\textbf{Takeaway.} The multilingual adaptation process does not inherently introduce regional or cultural biases, provided the base English classifier and embedding models are unbiased.
\end{block}

\begin{figure}[!htb]
    \centering
    \includegraphics[width=\linewidth]{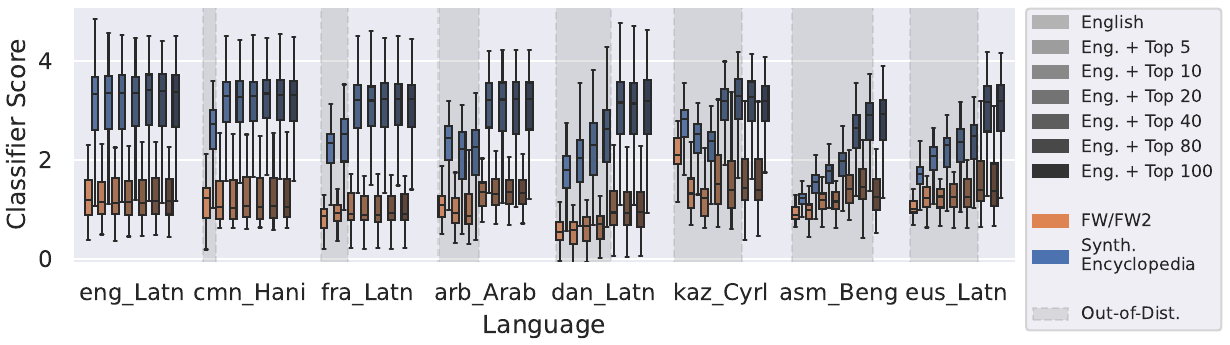}
    \caption{Comparison of quality scores between our mFW-edu classifier with mmBERT embeddings across selected languages-script pairs (ISO 639-3) among the top 100, evaluated on FineWeb (FW), FineWeb 2 (FW2), and synthetic encyclopedia data.}
    \label{fig:generalization_top100}
\end{figure}

\paragraph{Does the Classifier Generalize Across Languages?} To evaluate cross-lingual generalization of our approach, we compare mFW-edu classifier with mmBERT embeddings scores on FineWeb and FineWeb~2 data against our synthetic encyclopedia data. We test the classifier across a diverse selection of scripts and language families, trained on subsets of English with languages from top 5 to top 100, allowing us to characterize both in-distribution and out-of-distribution behaviour. We present the results in Figure~\ref{fig:generalization_top100}. We see that for in-distribution languages, scores for web and synthetic data align closely. In out-of-distribution cases, the classifier successfully differentiates between the two data distributions, although with lower score values, with performance improving as more training languages are added. Notably, these patterns hold even for the language isolate Basque (\texttt{eus\_Latn}), suggesting that the classifier effectively leverages the underlying embedding model to generalize beyond its explicit training set.

\begin{block}
\textbf{Takeaway.} The multilingual adaptation ensures consistent quality scoring across languages. While absolute scores for out-of-distribution languages may differ from in-distribution scores, the classifier maintains its ability to differentiate text quality.
\end{block}

\begin{figure}[!htb]
    \centering
    \includegraphics[width=\linewidth]{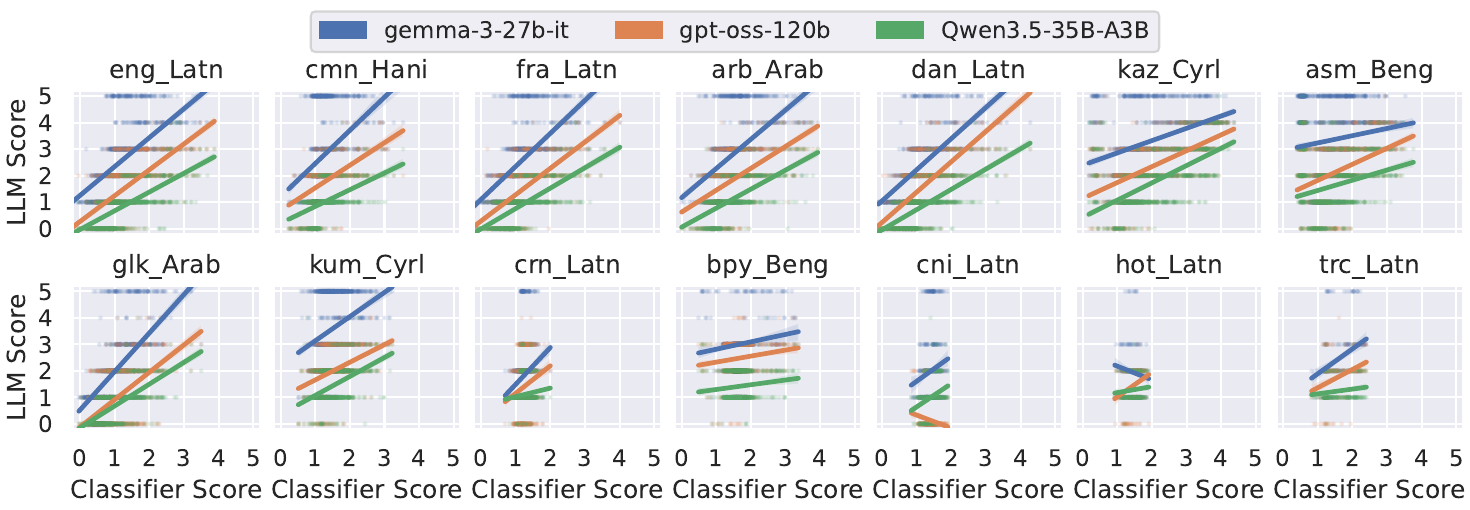}
    \caption{Comparison of our mFW-edu classifier with mmBERT embeddings against LLM-based scores (\texttt{gemma-3-27b-it}, \texttt{gpt-oss-120b}, \texttt{Qwen3.5-35B-A3B}). Top row: in-distribution (top 100), bottom row: out-of-distribution language-script pairs (beyond top 100; ISO 639-3).}
    \label{fig:generalization_all}
\end{figure}

To evaluate how well our findings generalize beyond the top 100 languages, we measure the correlation between our mFW-edu classifier with mmBERT embeddings and LLM-based judges using the FineWed-edu-like prompt. We use \texttt{gemma-3-27b-it}, \texttt{gpt-oss-120b}, and \texttt{Qwen3.5-35B-A3B} as judges to ensure cross-model alignment. We present the results in Figure~\ref{fig:generalization_all}. We see a consistent positive correlation for both in-distribution (top row; top 100 languages) and out-of-distribution languages (bottom row; beyond top 100 languages), with the only exceptions being \texttt{cni\_Latn} for \texttt{gpt-oss-120b} and \texttt{hot\_Latn} for \texttt{gemma-3-27b-it}, suggesting the classifier successfully captures the original scoring criteria beyond its initial language training distribution.

\begin{block}
\textbf{Takeaway.} Multilingually-adapted English classifiers can generalize the scoring criteria beyond the languages included in their training.
\end{block}

\section{Conclusion}

In this work, we presented a simple and scalable approach for training multilingual quality classifiers derived from English ones. Our experiments demonstrate that these classifiers perform on par with native multilingual baselines without introducing English-centric biases. Furthermore, we showed that by leveraging multilingual encoder-only embeddings and machine-translated text, these models generalize quality criteria to languages outside the training distribution, although we note a shift in score values for out-of-distribution languages. Importantly, this approach provides a viable path toward universal text quality classifiers by using research in the high-resource language space to support long-tail languages. Additionally, an interesting future research direction is exploring the application of this approach to different filter types, such as toxicity, as it could broaden the LLM democratization to global environments.

\newpage %

\bibliography{colm2026_conference}
\bibliographystyle{colm2026_conference}

\newpage %

\appendix

\section{Translation Prompt Used for Multilingual Document Translation}\label{appendix:translation}

\begin{tcolorbox}[nobeforeafter, title=Translation prompt]
Translate the following text into English. You must preserve all stylistic characteristics, such as including using the appropriate version of proper nouns and maintaining the style of the original text. You must preserve all formatting characteristics, such as keeping the newlines, tabs, dashes, and typos of the original text. Do not summarize the text and do not add any additional text. Output the full translation, without skipping any sentences.

Translate the following text to English:
\\
\\
\texttt{[DOCUMENT TEXT]}
\end{tcolorbox}

\begin{tcolorbox}[
  title={Example of translated text using \texttt{Qwen3-32B}},
  skin=bicolor,
  colback=gray!5!white,
  colbacklower=gray!20!white
  ]
\textbf{Original Text (Italian).} 
WhatsApp, l’applicazione di messaggistica istantanea, continua a rinnovarsi velocemente, oggi con una nuova funzione.

In questo caso, la piattaforma consentirà ai suoi utenti di pixelare le foto prima di inviarle ai nostri contatti, ovvero possiamo “sfocare” un volto che non vogliamo che appaia.

Nel sistema operativo iOS questa possibilità era disponibile diversi mesi fa, ma ora verrà aggiunta ai dispositivi Android, che avranno anche questo strumento utile e divertente per lavorare con le proprie immagini all’interno dell’app.

Come possiamo pixelare una foto con WhatsApp

-Tocca il pulsante allega e poi su “foto e video”.

-Scegli l’immagine dalla galleria.

-Tocca la matita nell’angolo in alto a destra e scegli il “pixelato” nella barra dei colori, cioè sotto tutto. Può anche trovarsi nella parte inferiore dello schermo in alcuni casi.

-Passa il dito sull’area che vuoi sfocare.

-Invia e fatto.

\tcblower

\textbf{Translated Text (English).} 
WhatsApp, the instant messaging application, continues to quickly renew itself, today with a new feature.

In this case, the platform will allow its users to pixelate photos before sending them to our contacts, that is, we can "blur" a face we don't want to appear.

On the iOS operating system this possibility was available several months ago, but now it will be added to Android devices, which will also have this useful and fun tool to work with their images within the app.

How can we pixelate a photo with WhatsApp

-Tap the attach button and then on "photo and video".

-Choose the image from the gallery.

-Tap the pencil in the top right corner and choose "pixelated" on the color bar, that is, at the bottom. It may also be found at the bottom of the screen in some cases.

-Swipe your finger over the area you want to blur.

-Send and done.
\end{tcolorbox}

\newpage %

\section{Comparison Between English and Multilingually-Adapted Quality Classifiers}\label{appendix:eng_multiling_classifier_comparison}

\begin{figure}[!htb]
  \begin{subfigure}{\textwidth}
    \includegraphics[width=\linewidth]{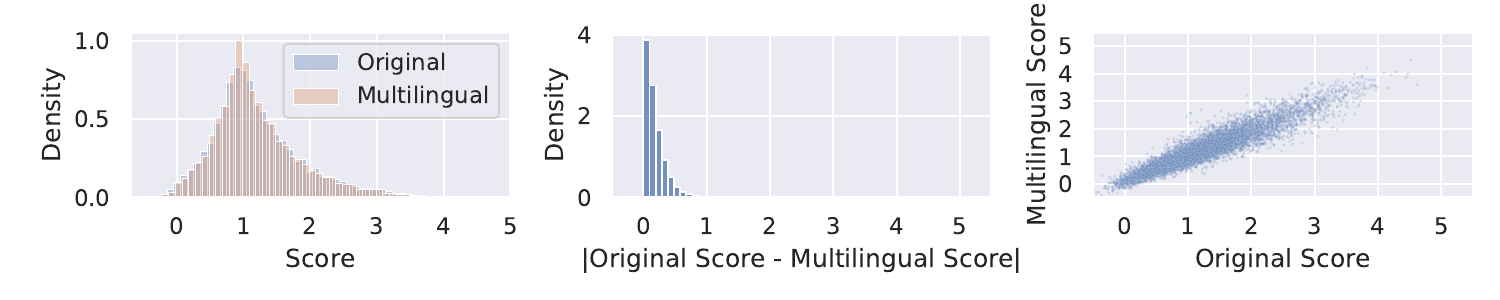}
    \caption{FineWeb-edu classifier with mmBERT embeddings}
  \end{subfigure}
  \\
  \begin{subfigure}{\textwidth}
    \includegraphics[width=\linewidth]{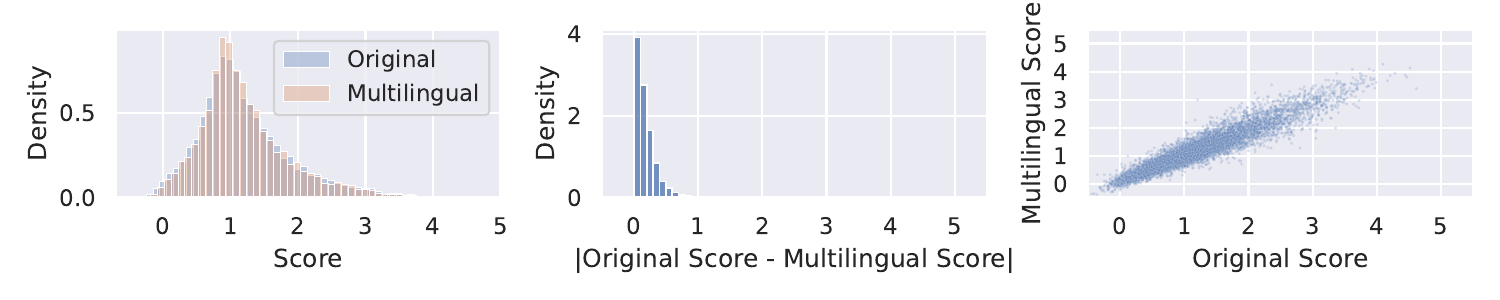}
    \caption{FineWeb-edu classifier with XLM-RoBERTa embeddings}
  \end{subfigure}
  \\
  \begin{subfigure}{\textwidth}
    \includegraphics[width=\linewidth]{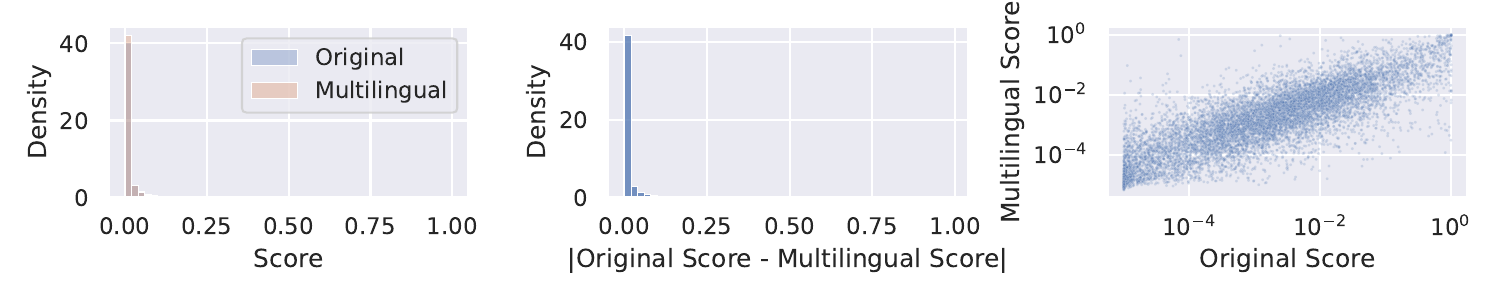}
    \caption{DCLM classifier with mmBERT embeddings}
  \end{subfigure}
  \begin{subfigure}{\textwidth}
    \includegraphics[width=\linewidth]{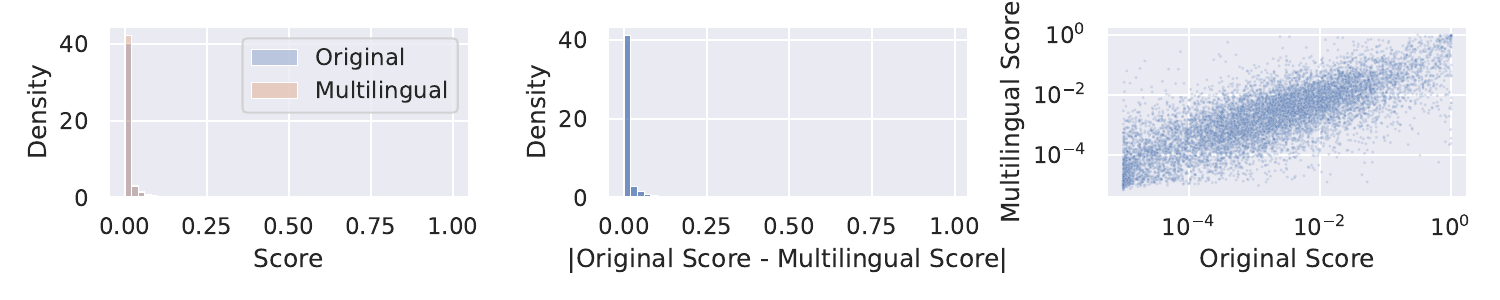}
    \caption{DCLM classifier with XLM-RoBERTa embeddings}
  \end{subfigure}
\caption{Score distribution of the original classifiers and the multilingual adaptations (left), score error distribution (middle), and the correlation between between their scores (right) on English FineWeb data.} \label{fig:eng_multiling_classifier_comparison}
\end{figure}

Figure~\ref{fig:eng_multiling_classifier_comparison} shows the score distribution of the original classifiers and the multilingual adaptations, score error distribution, and the correlation between between their scores on English FineWeb data.

\section{Configuration of the 1B, 3B, and 8B Paramter LLMs}\label{appendix:pretrain_configs}

In our experiments, we use the following training configurations for the LLMs:
\begin{itemize}
    \item Apertus-based 1B parameter multilingual models trained from scratch on 100B tokens of data in English and top 20 languages, with a batch size of  $\sim$2M tokens, a peak learning rate of 1.5e-4, 4\% linear warmup steps and 20\% 1-sqrt decay steps.
    \item Apertus-based 1B monolingual models trained from scratch on 30B tokens of data for several languages, with a batch size of $\sim$2M tokens, a peak learning rate of 1.5e-4, $\sim$6\% linear warmup steps and 20\% 1-sqrt decay steps.
    \item Apertus-based 3B parameter model cooled down on 50B tokens of data in all FineWeb and FineWeb~2 languages starting from a 200B token pretrained checkpoint trained on a mixture of 85\% FineWeb and 15\% top-33\% FineWeb2-HQ data from Apertus pretraining phase 1~\citep{apertus2025apertusdemocratizingopencompliant}, with a batch size of $\sim$2M tokens, from a peak learning rate of 1.5e-4, using 1-sqrt decay.
    \item Apertus 8B parameter model cooled down on 63B tokens of data in all FineWeb and FineWeb~2 languages starting from a 9.5T token pretrained Apertus checkpoint~\citep{apertus2025apertusdemocratizingopencompliant}, with a batch size of $\sim$4M tokens, from a peak learning rate of 1.1e-4, using 1-sqrt decay.
\end{itemize}

\section{Synthetic Encyclopedia Data Generation Process}\label{appendix:synth_encyclopedia}

We manually define 19 diverse categories and 4 seed article titles for each category. Then, we prompt \texttt{Qwen3-8B}\footnote{\url{https://huggingface.co/Qwen/Qwen3-8B}} to generate 20 diverse article titles based on the defined category and its 4 seed article titles. This gives us the following list of categories and article titles:
\begin{itemize}
    \item \textbf{Monuments}: Great Wall of China, Pyramids of Giza, Taj Mahal, Chichen Itza, Petra, Machu Picchu, Angkor Wat, Santorini Sunken City, Stonehenge, The Colosseum, Great Mosque of Djenné, Sagrada Família, Taj Mahal, Kailash Temple Complex, Lighthouse of Alexandria, Statue of Unity, Burj Khalifa, Sydney Opera House, Hagia Sophia, Leaning Tower of Pisa
    \item \textbf{Scientists}: Marie Curie, Nikola Tesla, Ada Lovelace, Alan Turing, Rosalind Franklin, Carl Sagan, Jane Goodall, Stephen Hawking, Albert Einstein, Gregor Mendel, Isaac Newton, Galileo Galilei, Henrietta Swan Leavitt, Lise Meitner, Katherine Johnson, William Herschel, Barbara McClintock, James Clerk Maxwell, Subrahmanyan Chandrasekhar, Albert Hofmann
    \item \textbf{Philosophers}: Aristotle, Socrates, Confucius, Lao Tzu, Plato, Marcus Aurelius, Ibn Sina, Immanuel Kant, Buddha, David Hume, Simone de Beauvoir, John Stuart Mill, Friedrich Nietzsche, Epictetus, Mencius, Karl Marx, Simone Weil, Bertrand Russell, Greta Garbo, Jean-Paul Sartre
    \item \textbf{Games}: The Legend of Zelda, Minecraft, Fortnite, Call of Duty, Mario Kart, Animal Crossing, Apex Legends, Tetris, SimCity, Dark Souls, Stardew Valley, Diablo, Pokémon Red and Blue, Portal, League of Legends, Rocket League, Red Dead Redemption 2, The Sims, Halo, Grand Theft Auto
    \item \textbf{Cities}: San Francisco, Beijing, Paris, Tokyo, Rio de Janeiro, Dubai, Cape Town, Sydney, Moscow, New York City, Mumbai, Barcelona, Oslo, Singapore, Santiago, Bangkok, Johannesburg, Lisbon, Auckland, Cape Town
    \item \textbf{Tech companies}: Alphabet, Apple, Amazon, Microsoft, Alibaba Group, Tencent, Samsung Electronics, NVIDIA, SoftBank, Intel, IBM, Salesforce, Baidu, Xiaomi, Siemens, SAP, Oracle, Cisco, LinkedIn, NVIDIA
    \item \textbf{Plants}: rose, birch, bamboo, orchid, cactus, maple, fern, jasmine, eucalyptus, lotus, succulent, magnolia, pine, dandelion, lavender, azalea, fern, sunflower, hydrangea, ivy
    \item \textbf{Animals}: horse, armadillo, elephant, kangaroo, penguin, giraffe, sloth, crocodile, tiger, zebra, octopus, dolphin, wolf, bear, snake, monkey, parrot, whale, rhinoceros, lion
    \item \textbf{Countries}: Japan, Mexico, Canada, Brazil, Australia, Norway, Kenya, Peru, South Africa, Saudi Arabia, India, Chile, Iceland, Nigeria, Thailand, Finland, Colombia, Indonesia, Spain, Argentina
    \item \textbf{Mythology}: Anansi, Athena, Amaterasu, Odin, Yama, Maui, Anubis, Tiamat, Quetzalcoatl, Cernunnos, Amaterasu, Shangdi, Ra, Indra, Tlaloc, Njord, Balder, Thoth, Anansi, Loki
    \item \textbf{Literature}: Shakespeare, Maya Angelou, Gabriel García Márquez, Leo Tolstoy, Franz Kafka, Chinua Achebe, Jorge Luis Borges, Sylvia Plath, Haruki Murakami, Toni Morrison, Anton Chekhov, Isabel Allende, Fyodor Dostoevsky, Gabriel García Márquez, James Joyce, Salman Rushdie, Margaret Atwood, Paulo Coelho, Alice Walker, J.K. Rowling
    \item \textbf{Movies}: The Godfather, Inception, Parasite, Amélie, Slumdog Millionaire, Roma, The Lives of Others, Tokyo Story, Life is Beautiful, Pan's Labyrinth, Once Upon a Time in Mexico, The Last Emperor, La Haine, The Pianist, Wong Kar-wai's Happy Together, Koyaanisqatsi, 12 Years a Slave, The Act of Killing, Birdman, Silver Linings Playbook
    \item \textbf{Music}: Elvis Presley, Queen, Bob Marley, The Beatles, Jimi Hendrix, Madonna, BTS, Coldplay, Kacey Musgraves, Bad Bunny, Tahiya Kariuki, Sia, Ed Sheeran, Beyoncé, Shakin' Stevens, Rammstein, Shaggy, Ani DiFranco, Shinedown, Lady Gaga
    \item \textbf{Art Movements}: Surrealism, Mannerism, Art Deco, Naïve Art, Tonalism, Post-Impressionism, Art Nouveau, Fauvism, Dada, Ukiyo-e, Symbolism, Kinetic Art, Pop Art, Op Art, Minimalism, Expressionism, Nihonga, Abstract Expressionism, Constructivism, Afrofuturism
    \item \textbf{Historical Events}: World War II, French Revolution, Fall of the Berlin Wall, American Civil War, Russian Revolution, Treaty of Versailles, Spanish Inquisition, Industrial Revolution, Arab Spring, Cold War, Fall of the Soviet Union, Black Death, Mexican Revolution, D-Day Invasion, Roman Empire Decline, Hundred Years' War, Magna Carta, First World War, Great Depression, Moon Landing
    \item \textbf{Religions}: Christianity, Islam, Hinduism, Buddhism, Judaism, Sikhism, Taoism, Confucianism, Shinto, Jainism, Zoroastrianism, Baha'i Faith, Buddhism (Theravada), Buddhism (Mahayana), Sikhism, Taoism, Shinto, Jainism, Animism, Scientology
    \item \textbf{Sports}: Cricket, Rugby Union, Judo, Taekwondo, Ice Hockey, Volleyball, American Football, Baseball, Boxing, Cycling, Rugby Sevens, Handball, Badminton, Squash, Table Tennis, Football (Soccer), Fencing, Water Polo, Karate, Field Hockey
    \item \textbf{Symbols}: Olympic rings, American flag, Japanese rising sun, Egyptian ankh, Chinese dragon, Mexican national football team logo, Indian lotus, Australian outback, French tricolor, British Union Jack, Nigerian flag, South African rainbow, Canadian maple leaf, Swiss cross, Norwegian lynx, New Zealand kiwi, Brazilian flag, Russian double-headed eagle, Irish tricolor, Israeli Star of David
    \item \textbf{Natural Landmarks}: Grand Canyon, Victoria Falls, Great Barrier Reef, Angel Falls, Mount Fuji, Sahara Desert, Amazon Rainforest, Uluru, Denali, Mariana Trench, Northern Lights (Aurora Borealis), Patagonia, Great Barrier Reef, Iceland's Blue Lagoon, Kilimanjaro, Galápagos Islands, Dead Sea, Serengeti Plain, Victoria Falls, Niagara Falls
\end{itemize}

We use the following prompt to generate synthetic encyclopedia data using the \texttt{Qwen3.5-35B-A3B} model for English (\texttt{eng\_Latn}), Chinese (\texttt{cmn\_Hani}), French (\texttt{fra\_Latn}), Arabic (\texttt{arb\_Arab}), Danish (\texttt{dan\_Latn}), Kazakh (\texttt{kaz\_Cyrl}), Assamese (\texttt{asm\_Beng}), and Basque (\texttt{eus\_Latn}) languages.

\begin{tcolorbox}[nobeforeafter, title=Synthetic encyclopedia generation prompt]
Generate an encyclopedia-style article talking about \texttt{[ARTICLE TITLE]} in the context of \texttt{[CATEGORY]}. You should only output this without any other text. Your output must be in the \texttt{[LANGUAGE]} language with the ISO 639-3 langauge and script code \texttt{[LANGUAGE CODE AND SCRIPT]}.
\end{tcolorbox}

We show an example of a generated article.

\newpage %

\begin{tcolorbox}[nobeforeafter, title={Example of a synthetic encyclopedia article with the title \textit{Octopus}, in the category of \textit{Animals}, in French (\texttt{fra\_Latn})}]
\# Poulpe
\\
\\
Le poulpe (du latin *polypus*, et du grec *polypous*, « pied multiple ») est un céphalopode marin de la classe des Céphalopodes, ordre des Octopodes. Il se distingue par son corps mou, dépourvu de squelette interne (à l'exception de la radule chez certaines espèces), et par ses huit bras musclés, chacun pourvu de deux rangées de ventouses. Les poulpes appartiennent à l'embranchement des Mollusques, tout comme les escargots et les moules, mais ils en sont les parents les plus éloignés parmi les invertébrés marins.
\\
\\
\#\# Description physique
\\
\\
Le corps du poulpe, appelé manteau, est généralement de forme ovoïde ou sphérique. Il est composé de muscles lisses et striés qui lui permettent une grande flexibilité et la capacité de s'infiltrer dans des espaces extrêmement restreints. La tête du poulpe, située à l'avant du manteau, abrite le cerveau et les organes sensoriels.
\\
\\
Ses huit bras sont devenus l'élément caractéristique de l'ordre. Contrairement aux tentacules des autres céphalopodes comme les calamars, les bras des poulpes sont de longueur égale et couverts de ventouses sur toute leur longueur. Chaque ventouse possède sa propre capacité de mouvement et contient des chémorécepteurs, permettant au poulpe de « goûter » ce qu'il touche. Le cerveau est unique dans le règne animal pour sa distribution : bien que le cerveau central contrôle les fonctions globales, un ganglion nerveux existe à la base de chaque bras, permettant à ces derniers d'agir de manière autonome et de résoudre des problèmes complexes sans intervention immédiate du cerveau central.
\\
\\
\#\# Habitat et répartition
\\
\\
Les poulpes sont présents dans tous les océans du monde, du niveau de la mer jusqu'à des profondeurs abyssales dépassant les 6 000 mètres. On les trouve aussi bien dans les eaux tropicales des récifs coralliens que dans les eaux froides des pôles. La plupart des espèces sont benthiques, vivant au fond de l'océan, bien que certaines espèces, comme le poulpe volant (*Octopus volans*), puissent planer dans la colonne d'eau.
\\
\\
\#\# Alimentation et comportement
\\
\\
Les poulpes sont des prédateurs carnivores opportunistes. Leur régime alimentaire varie selon les espèces et comprend principalement des crustacés (crabes, crevettes), des mollusques (huîtres, palourdes) et des petits poissons. Ils utilisent leurs bras pour saisir leurs proies, les soumettre avec leur bec corné situé au centre des bras, et souvent pour ouvrir les coquilles, parfois en utilisant des pierres comme outils.

\texttt{[...]}
\end{tcolorbox}

\begin{figure}[!htb]
    \centering
    \includegraphics[width=\linewidth]{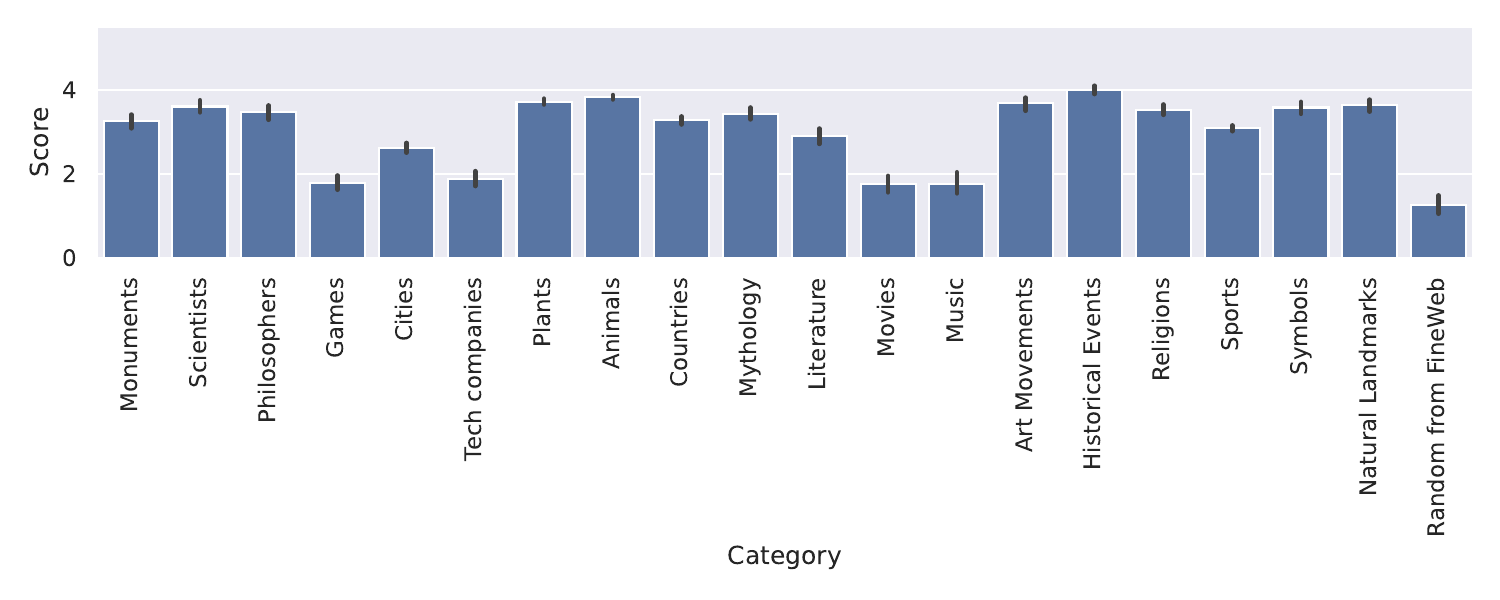}
    \caption{Score distribution of our multilingually adapted mFW-edu classifier with mmBERT embeddings on the English synthetic encyclopedia data.}
    \label{fig:synth_encyclopedia_mfwedu_scores}
\end{figure}

In Figure~\ref{fig:synth_encyclopedia_mfwedu_scores}, we show the score distribution of our multilingually adapted mFW-edu classifier with mmBERT embeddings on the English synthetic encyclopedia data.

\section{LLM-as-a-judge Prompt and Example}\label{appendix:llm_judge}

We use the original FineWeb-edu prompt and adapt it only by adding language information. When applicable, we disable the LLMs' reasoning outputs.

\begin{tcolorbox}[nobeforeafter, title=LLM-as-a-judge multilingual FineWeb-edu prompt]
Below is an extract from a web page in the language with ISO 639-3 language and script code: \texttt{[LANGUAGE CODE AND SCRIPT]}. Evaluate whether the page has a high educational value and could be useful in an educational setting for teaching from primary school to grade school levels using the additive 5-point scoring system described below. Points are accumulated based on the satisfaction of each criterion:
\\
\\
- Add 1 point if the extract provides some basic information relevant to educational topics, even if it includes some irrelevant or non-academic content like advertisements and promotional material.

- Add another point if the extract addresses certain elements pertinent to education but does not align closely with educational standards. It might mix educational content with non-educational material, offering a superficial overview of potentially useful topics, or presenting information in a disorganized manner and incoherent writing style.

- Award a third point if the extract is appropriate for educational use and introduces key concepts relevant to school curricula. It is coherent though it may not be comprehensive or could include some extraneous information. It may resemble an introductory section of a textbook or a basic tutorial that is suitable for learning but has notable limitations like treating concepts that are too complex for grade school students. 

- Grant a fourth point if the extract highly relevant and beneficial for educational purposes for a level not higher than grade school, exhibiting a clear and consistent writing style. It could be similar to a chapter from a textbook or a tutorial, offering substantial educational content, including exercises and solutions, with minimal irrelevant information, and the concepts aren't too advanced for grade school students. The content is coherent, focused, and valuable for structured learning.

- Bestow a fifth point if the extract is outstanding in its educational value, perfectly suited for teaching either at primary school or grade school. It follows detailed reasoning, the writing style is easy to follow and offers profound and thorough insights into the subject matter, devoid of any non-educational or complex content.
\\
\\
The extract:

\texttt{[DOCUMENT TEXT]}
\\
\\
After examining the extract: 

- Briefly justify your total score, up to 100 words.

- Conclude with the score using the format: "Educational score:  <total points>"
\end{tcolorbox}

\begin{tcolorbox}[
  title={LLM-as-a-judge example response of \texttt{Qwen3.5-35B-A3B}},
  skin=bicolor,
  colback=gray!5!white,
  colbacklower=gray!20!white
  ]
    \textbf{Original Text (Danish).}
Hvis du leder efter en busk, der kan blomstre i skygge, kan den yndefulde deutzia muligvis være planten for dig. Denne højeformede busk \& rsquo; s rigelige blomster og fleksible dyrkningsbetingelser er en klar plus for mange gartnere.

Hvad er Deutzia?

Deutzia er en gruppe på omkring 60 arter, hvoraf de fleste er oprindelige til Kina og andre steder i Asien, mens nogle få stammer fra Europa og Mellemamerika. Disse haugformende buske har lange, buede grene, der giver dem et grædende eller kaskaderende udseende.

Deutzias er medlemmer af hortensiafamilien og kan ligesom hortensiaer , de producerer små blomster, der vokser rigeligt i klynger. Deutzia-blomster ser dog ganske forskellige ud, med kronbladene af nogle arter langstrakte og forsigtigt hængende og andre klokkeformede eller åbne. Disse duftende blomster er rene hvide eller farvet med lyserødt, og de vises i cirka to uger i det tidlige forår til midten af sommeren.

Deutzia producerer løvfældende, lysegrønne blade, og nogle sorter udvikler røde blade om efteråret. Disse buske er pryd i vinterhalvåret med bark, der skrælner tilbage for at afsløre en rødlig-orange farve nedenunder.

\texttt{[...]}
    \tcblower
    \textbf{LLM Response.} The extract provides clear, factual information about the Deutzia plant, covering its origins, characteristics, care requirements, and specific varieties. The content is coherent and well-structured, resembling a basic tutorial or encyclopedia entry. It introduces key botanical concepts suitable for primary or grade school students, explaining terms like "zones" and "pruning" simply. However, it lacks interactive elements like exercises, and some details (e.g., specific USDA zones) might require slight contextualization for younger learners. It does not reach the level of a comprehensive textbook chapter with deep insights or perfect pedagogical design, but it is highly useful for general learning.
\\
\\
Educational score: 3
\end{tcolorbox}

\section{Impact of Including Varying Amount of Languages in Quality Classifier Training}\label{appendix:varying_languages_classifier}

Table~\ref{tab:full_classifier_training_languages_comparison} shows the full evaluation benchmark results when training our multilingually adapted mFW-edu classifier with mmBERT embeddings on a varying amount of languages, compared against our baselines.

\begin{table}[!htb]
    \centering
\resizebox{\textwidth}{!}{

\begin{tabular}{lcccccccccc}
\toprule[2pt]
& \textbf{GMMLU$_c$} & \textbf{INCLUDE$_c$} & \textbf{ARC} & \textbf{mARC} & \textbf{HS} & \textbf{mHS} & \textbf{XNLI} & \textbf{OBQA} & \textbf{XWG} & \textbf{Avg.} \\
\textbf{Model} &  &  &  &  &  &  &  &  &  & \\

\midrule[2pt]
\multicolumn{11}{l}{\textit{1B Model, 100B Tokens from Scratch, Training and Evaluation on English + Top 20 Languages}} \\
\midrule

FW+FW2 & 0.2743 & 0.2817 & 0.3880 & 0.2586 & 0.5434 & 0.3887 & 0.4086 & 0.3180 & 0.6051 & 0.3852 \\
FWHQ+FW2HQ & 0.2893 & 0.2870 & 0.4461 & 0.2747 & 0.5641 & 0.4035 & \textbf{0.4214} & 0.3680 & \textbf{0.6308} & 0.4094 \\
FWHQ+FW2HQ$^{+}$ & 0.2952 & 0.2959 & 0.4505 & 0.2816 & \textbf{0.5762} & \textbf{0.4120} & 0.4116 & 0.3640 & 0.6221 & 0.4121 \\
mFW-edu$_{\text{Eng}}$ & 0.2925 & 0.2954 & 0.4766 & 0.2880 & 0.5275 & 0.3999 & 0.4202 & 0.3680 & 0.6022 & 0.4078 \\
mFW-edu$_{\text{Eng + Top 5}}$ & \textbf{0.3020} & 0.2952 & 0.4786 & 0.2948 & 0.5406 & 0.4007 & 0.4173 & 0.3740 & 0.6045 & 0.4120 \\
mFW-edu$_{\text{Eng + Top 10}}$ & 0.2852 & 0.2968 & 0.4813 & \textbf{0.2948} & 0.5420 & 0.3999 & 0.4054 & 0.3740 & 0.5936 & 0.4081 \\
mFW-edu$_{\text{Eng + Top 20}}$ & 0.2884 & 0.2966 & 0.4872 & 0.2932 & 0.5378 & 0.4022 & 0.4132 & \textbf{0.3940} & 0.6161 & \textbf{0.4143} \\
mFW-edu$_{\text{Eng + Top 40}}$ & 0.2880 & \textbf{0.2998} & 0.4819 & 0.2921 & 0.5384 & 0.4030 & 0.4155 & 0.3840 & 0.6065 & 0.4121 \\
mFW-edu$_{\text{Eng + Top 100}}$ & 0.2989 & 0.2913 & \textbf{0.4897} & 0.2937 & 0.5469 & 0.4026 & 0.4121 & 0.3740 & 0.6180 & 0.4141 \\

\bottomrule
\end{tabular}

}
    \caption{Benchmark performance of 1B parameter multilingual LLMs trained on 100B tokens. We compare models trained on data selected using our multilingually adapted mFW-edu classifier paired with mmBERT embeddings trained on a varying amount of languages, and the FW+FW2, FWHQ+FW2HQ, and FWHQ+FW2HQ$^{+}$ baselines. We retain the top 10\% of documents based on the classifier scores, except for FW+FW2 which is not model filtered.}
    \label{tab:full_classifier_training_languages_comparison}
\end{table}

\section{Disclosure of LLM Use}

In addition to the methodology described in the paper, we use LLMs to improve the clarity and phrasing of the paragraphs in the paper. We \textbf{do not} use LLMs to generate research ideas, new paragraphs or references in the paper.

\end{document}